\documentclass[letterpaper, 10 pt, conference]{ieeeconf}  

\IEEEoverridecommandlockouts                              

\usepackage{amsmath} 
\usepackage{amssymb}  

\usepackage{wrapfig}
\usepackage{graphicx}
\usepackage{booktabs} 
\usepackage{color}
\usepackage{cuted}
\usepackage{caption}
\usepackage{hyperref}
\usepackage{algorithm}
\usepackage{algpseudocode}
\usepackage{listings}
\usepackage{float}
\usepackage[dvipsnames]{xcolor}
\usepackage{hyperref}
\usepackage{multirow}
\usepackage[numbers,sort&compress]{natbib}
\usepackage{mdframed}
\usepackage{adjustbox}

\hypersetup{
    colorlinks=true,
    linkcolor=Blue,
    filecolor=magenta,
    citecolor=Blue,
    urlcolor=Blue,
    pdftitle={Overleaf Example},
    pdfpagemode=FullScreen,
    }

\title{\LARGE \bf
    KUDA: Keypoints to Unify Dynamics Learning and Visual Prompting for Open-Vocabulary Robotic Manipulation
}

\author{Zixian Liu$^{1*}$, Mingtong Zhang$^{2*}$, and Yunzhu Li$^{3}$
\thanks{$^*$denotes equal contribution. $^{1}$Tsinghua University, $^{2}$University of Illinois Urbana-Champaign, $^{3}$Columbia University}
}

\begin{document}

\maketitle
\thispagestyle{empty}
\pagestyle{empty}


\begin{strip}
    \centering
    \vspace{-56pt}
    \includegraphics[width=\linewidth]{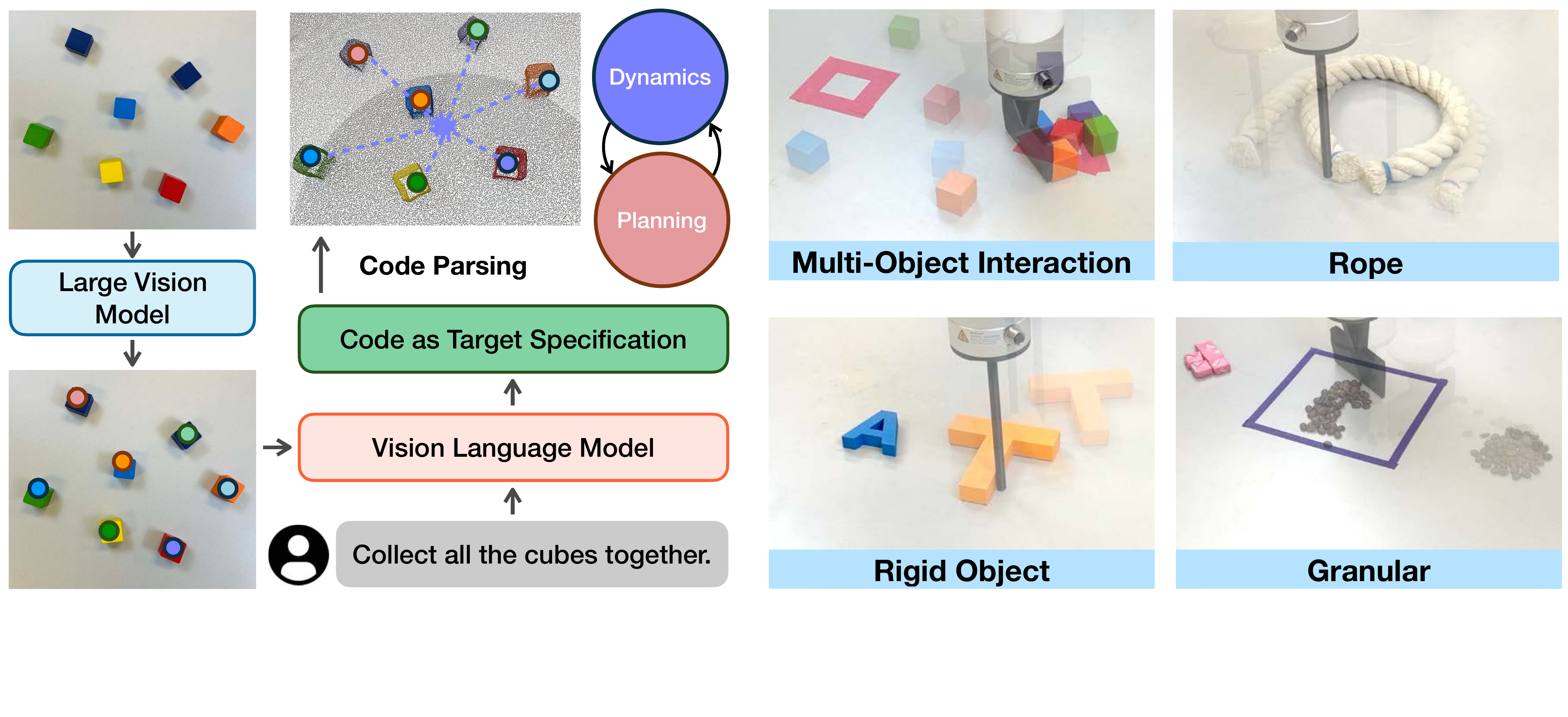}
    \vspace{-55pt}
    \captionof{figure}
    {\small
    \textbf{KUDA} is an open-vocabulary manipulation system that uses keypoints to unify the visual prompting of vision language models (VLMs) and dynamics modeling. Taking the RGBD observation and the language instruction as inputs, KUDA samples keypoints in the environment, then uses a VLM to generate code specifying keypoint-based target specification. These keypoints are translated into a cost function for model-based planning with learned dynamics models, enabling open-vocabulary manipulation across various object categories.
    }
    \vspace{-10pt}
    \label{fig:teaser}
\end{strip}

\begin{abstract}


With the rapid advancement of large language models (LLMs) and vision-language models (VLMs), significant progress has been made in developing open-vocabulary robotic manipulation systems. However, many existing approaches overlook the importance of object dynamics, limiting their applicability to more complex, dynamic tasks. In this work, we introduce KUDA, an open-vocabulary manipulation system that integrates dynamics learning and visual prompting through keypoints, leveraging both VLMs and learning-based neural dynamics models. Our key insight is that a \textit{keypoint-based target specification} is simultaneously interpretable by VLMs and can be efficiently translated into cost functions for model-based planning. Given language instructions and visual observations, KUDA first assigns keypoints to the RGB image and queries the VLM to generate target specifications. These abstract keypoint-based representations are then converted into cost functions, which are optimized using a learned dynamics model to produce robotic trajectories. We evaluate KUDA on a range of manipulation tasks, including free-form language instructions across diverse object categories, multi-object interactions, and deformable or granular objects, demonstrating the effectiveness of our framework. The project page is available at \url{http://kuda-dynamics.github.io}.

\end{abstract}


\section{Introduction}



It has been a longstanding focus to create an open-vocabulary robotic system capable of executing tasks based on human language in diverse environments.
However, human language is inherently abstract and often ambiguous, requiring contextual knowledge and the ability to ground language inputs in the environments where the robots operate.
Recent advances in large language models (LLMs) and vision-language models (VLMs)~\cite{Devlin2019BERTPO, radford2018improving, radford2019language, brown2020language, chowdhery2023palm, achiam2023gpt, radford2021learning, li2022blip, li2023blip} have demonstrated advanced capabilities in text and image understanding. These developments have paved new paths for incorporating such models into robotic systems~\cite{huang2023voxposer, liu2024moka, huang2024rekep}.
However, many existing open-vocabulary robotic systems heavily rely on VLMs and LLMs for guidance at all levels and do not provide an explicit account of object dynamics. As a result, they typically focus only on rigid objects and coarse-grained manipulation, limiting their applicability to more complex and dynamic tasks involving diverse object categories, such as deformable objects and object piles.

On the other hand, learning-based dynamics models have shown the ability to model the complex behaviors of real-world objects directly from observation data~\cite{finn2017deep, nagabandi2020deep, battaglia2016interaction, li2018learning}. These models can accurately predict the future states of objects with varying object categories and shapes, accounting for different interactions. However, model-based planning with dynamics models typically requires a pre-defined target state or cost function, which cannot be directly inferred from high-level language instructions. This raises a key question: \textit{How can we develop open-vocabulary manipulation systems that harness the flexibility of task specification via VLMs while preserving the benefits of model-based planning?}

Our key insight is to develop a unified keypoint representation that integrates dynamics learning and visual prompting through VLMs. Using keypoints as the visual representation is intuitive for vision-language models to interpret and express, while also being precisely defined and easily translatable into a cost function for planning with dynamics models. To achieve this, we propose defining the objective function for planning using keypoints and employing mark-based visual prompting, inspired by~\cite{liu2024moka}, to enable the VLM to generate code that specifies the objective as arithmetic relationships between the visual keypoints.

Building on our keypoint-based target specifications, we introduce \textbf{KUDA}: an open-vocabulary manipulation system that utilizes \textbf{K}eypoints to \textbf{U}nify \textbf{D}ynamics learning and visu\textbf{A}l prompting. KUDA employs an upstream VLM along with a pre-trained dynamics model, using keypoints as a shared intermediate representation. Given a language instruction for the manipulation task and the current visual observation of the experimental setup, KUDA automatically samples and labels keypoints from the RGB image. The VLM is then prompted to generate target specifications, which are subsequently converted into a cost function. During robot execution, a two-level closed-loop control mechanism ensures effective and robust model-based planning. Notably, we found that incorporating few-shot examples significantly enhances the performance of the VLM. Inspired by this, we developed a prompt library and a retrieval mechanism based on score matching, ensuring high-quality few-shot examples without exceeding input token limits.
In summary, our contributions are as follows:

\begin{itemize}
    \item We propose using keypoints as a unified intermediate representation to bridge dynamics learning and visual prompting through VLMs.
    \item We design a prompt retriever that automatically subsamples from our prompt library based on the task description, while ensuring it stays within the context window of the VLMs.
    \item We evaluate the integrated system in real-world manipulation tasks, demonstrating state-of-the-art performance on tasks involving diverse object materials, such as ropes and granular objects (Fig.~\ref{fig:teaser}), and covering a range of language instructions.
\end{itemize}

\section{Related Work}

\subsection{Grounding Language Instructions}

Using human language to instruct intelligent robots has been an active research domain. However, most works~\cite{karamcheti2020learning, myers2024policy, ren2023leveraging} mainly focus on decomposing high-level language instructions as subtasks. Grounding ambiguous human language into structured action sequences that robots can execute remains a significant challenge~\cite{5453186, Misra-RSS-14, saycan2022arxiv}. Most existing approaches use action primitives as the basic elements for planning. These methods either employ classical techniques such as lexical analysis, formal logic, and graphical models~\cite{thomason2015learning, 5453186, tellex2011understanding, kollar2014grounding}, or leverage pre-trained large models to comprehend instructions and generate task plans~\cite{saycan2022arxiv, jiang2024roboexp, hu2023look, codeaspolicies2022}. However, the reliance on pre-defined motion primitives is often seen as a major limitation for developing a universal manipulation system~\cite{huang2023voxposer}.

Many recent works have also focused on grounding language into lower-level actions. These approaches include language-conditioned imitation learning~\cite{shridhar2021cliport, shridhar2022peract, jang2022bc}, synthesizing value maps or reward functions~\cite{sharma2022correcting, huang2023voxposer, yu2023language, huang2024rekep}, and generating motions through visual prompting~\cite{liu2024moka, nasiriany2024pivot}.
In our work, we propose a new paradigm to ground natural language instructions into keypoint target configurations for model-based planning with learned dynamics models.

\subsection{Foundation Models in Robotics}

The remarkable success of large language models (LLMs) has recently earned significant interest in the field of robotics. Recent works have explored how to effectively integrate these powerful models into robotic systems. Some approaches leverage LLMs for high-level task planning~\cite{chen2023nl2tl, chen2023autotamp}, while others have demonstrated that LLMs excel at generating code for robot control~\cite{codeaspolicies2022, 10161317, vemprala2024chatgpt}. Another application of LLMs is in synthesizing value functions or reward functions~\cite{huang2023voxposer, yu2023language, huang2022inner}. However, these approaches typically require converting both the task and observations into textual form, which often leads to suboptimal performance in real-world manipulation scenarios.


Recently, vision-language models (VLMs) have earned significant attention. In addition to replacing previous LLMs with powerful VLMs, such as GPT-4V~\cite{achiam2023gpt}, to achieve better grounding in real-world scenarios, VLMs have been applied in various contexts. Some pre-trained VLMs have shown superior perception capabilities~\cite{minderer2022simple, liu2023grounding, kirillov2023segment}, while others have been used to generate affordances or constraints based on visual prompts~\cite{liu2024moka, huang2024copa, huang2024rekep}. However, despite their capabilities, these advanced VLMs lack a good understanding of the 3D spatial relationships and object dynamics, making them inefficient in more complicated manipulation tasks.


Another recent trend focuses on collecting large-scale robotics data and training end-to-end general-purpose models for robotic tasks~\cite{rt12022arxiv, rt22023arxiv, open_x_embodiment_rt_x_2023, khazatsky2024droid}. However, many works in this direction focus primarily on rigid object manipulation and cannot perform complex manipulation across various object categories due to a lack of knowledge of object dynamics. Additionally, collecting data and training models for such works is highly resource-intensive, and their performance remains limited.
In our work, we leverage the state-of-the-art VLM to generate the target specifications for the neural dynamics models, which enables a flexible and robust framework that facilitates model-based planning.


\subsection{Visual Prompting and In-Context Learning}

Visual prompting is an emerging technique that has earned much attention with the development of large vision-language models. This technique enhances the visual grounding capabilities of these foundation models through incorporating visual markers into observations. Yang et al.~\cite{yang2023set} demonstrated that overlaying the original image with its semantic segmentation enables GPT-4V to answer questions requiring visual grounding. Cai et al.~\cite{cai2023making} introduced ViP-LLaVA, a model capable of decoding visual prompts that include various types of visual markers. Others have also been using visual prompts generated by predefined rules to produce affordances for robotic planning~\cite{liu2024moka, nasiriany2024pivot}.


In-context learning is a relatively novel paradigm in the vision domain. It allows a pre-trained model to efficiently adapt to a novel downstream task using a few input-output examples, without requiring fine-tuning. One of the earliest works in this area is Flamingo~\cite{alayrac2022flamingo}, a vision-language model (VLM) that is capable of learning from few-shot examples. Bar et al.~\cite{bar2022visual} approach the image-to-image in-context learning problem as an image inpainting task, while Yang et al.~\cite{yang2024imagebrush} introduce in-context learning for image editing. Li et al.~\cite{li2024visual} offer a general method for in-context learning in segmentation tasks. Additionally, Zhang et al.~\cite{zhang2023makes} provide a comprehensive study on how to select effective examples for visual in-context learning.
In our work, we utilize a set of well-designed prompting formulations to guide the vision language model in generating keypoint-based target specifications. Additionally, inspired by \cite{zha2024distilling}, we introduce a prompt retriever to ensure the quality of few-shot examples.



\begin{figure*}[thbp]
    \centering
    \includegraphics[width=\linewidth]{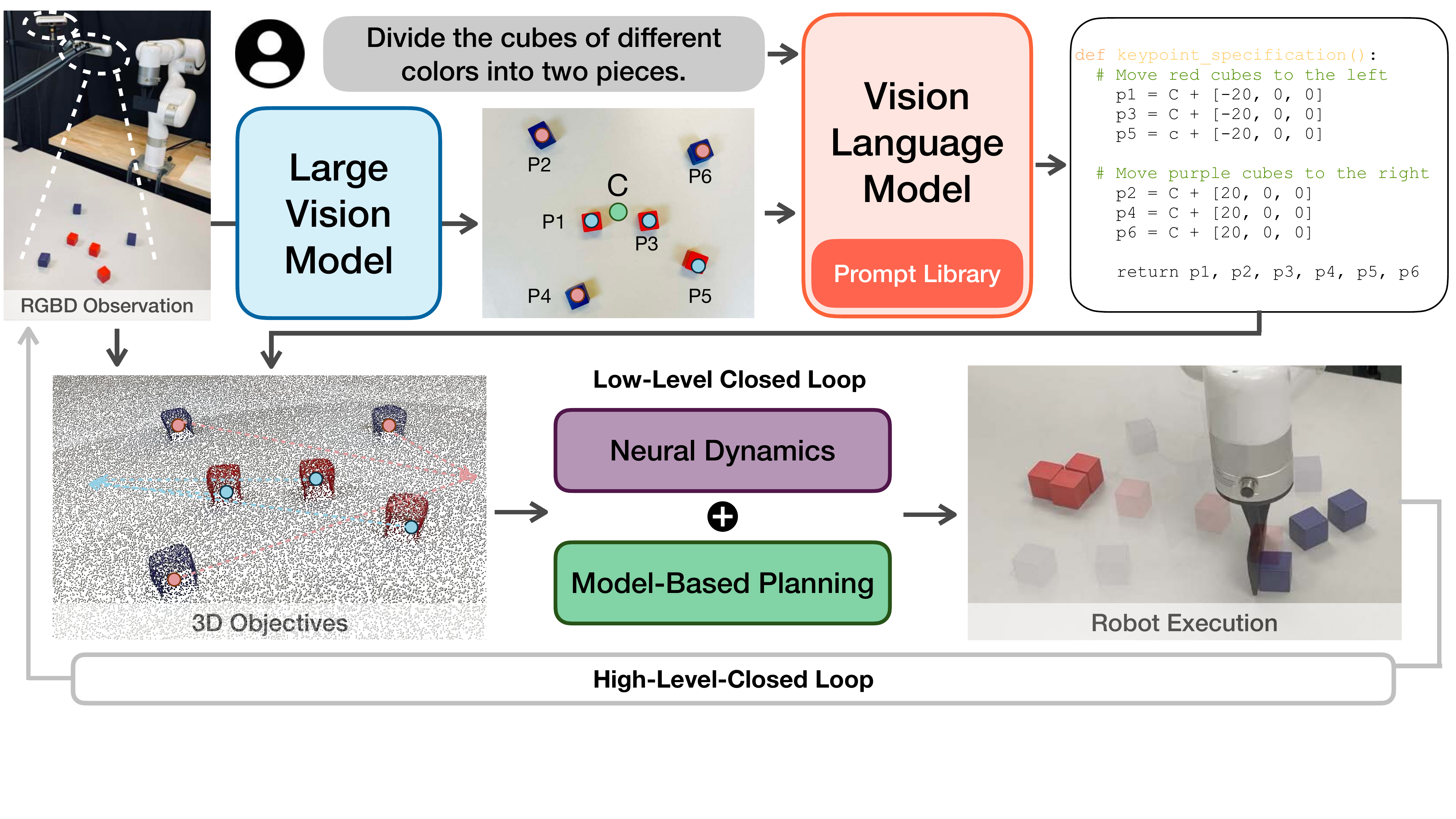}
    \vspace{-55pt}
    \caption{\small \textbf{Overview of KUDA}. Taking the RGBD observations and a language instruction as inputs, we first utilize the large vision model to obtain the keypoints and label them on the RGB image to obtain the visual prompt (green dot C marks the center reference point). Next, the vision-language model generates code for target specifications, which are projected into 3D space to construct the 3D objectives. Lastly, we utilize the pre-trained dynamics model for model-based planning. After a certain number of actions, the VLM is re-queried with the current observation, enabling high-level closed-loop planning to correct VLM and execution errors.}
    \label{fig:method}
    \vspace{-15pt}
\end{figure*}

\section{Method}

We first provide the problem formulation. Then we describe how we prompt the VLM to generate the keypoint-based \textit{target specification}. Next, we present the design of the prompt retriever and the Top-K prompt library. Lastly, we specify how we obtain the cost function from target specification and how we perform model-based planning with neural dynamics models, along with the two-level closed-loop control mechanism. The overview of our KUDA framework is shown in Fig.~\ref{fig:method}.

\subsection{Problem Formulation} \label{sec:3.1}

We consider a tabletop manipulation problem defined by a \textit{free-form} language instruction $\mathcal{L}$ that pertains to one or multiple objects $\mathcal{O}$, along with a pre-trained neural dynamics model $f$. The goal is to optimize the robot's action trajectory $\tau$ to perform the manipulation task described by $\mathcal{L}$. The instruction $\mathcal{L}$ is typically abstract and requires interpretation that incorporates both common sense and the context of the experiment configurations. The dynamics model $f$ is able to predict how the object set $\mathcal{O}$ will change when a specific action is applied:
\begin{equation}
\hat{z}_{t+1} = f(z_t, u_t),
\end{equation}
where $z_t$ and $z_{t+1}$ represent the current state of the environment at time $t$ and the subsequent state at time $t + 1$, respectively. $\hat{z}_{t+1}$ is the predicted state at time $t + 1$, and $u_t$ is the applied action. The environment state describes the positions of the 3D points ${o_i = (x_i, y_i, z_i)}$ that represent the objects (or parts of an object) in $\mathcal{O}$.


The key problem here is how to define the objectives from the ambiguous instruction $\mathcal{L}$ for optimization. We leverage VLMs to propose a keypoint-based representation, where we will obtain our cost function. Once the cost function $C(\cdot|\mathcal{L})$ is obtained from the instruction $\mathcal{L}$, the problem can be formulated as an optimization problem over the robot's action trajectory $\tau$:
\begin{equation}
\tau = \arg\min_\tau{ C(z'|\mathcal{L}) }, \label{1}
\end{equation}
where $z'$ represents the final state of the environment after the action sequence $\tau$ has been applied.


\subsection{Keypoint-Based Target Specification} \label{sec:3.2}

Inspired by \cite{liu2024moka}, we employ visual markers to enhance the visual grounding capabilities of vision-language models (VLMs). A key insight in our framework is that a wide range of manipulation tasks can be effectively described by the spatial relationships between keypoints on the objects to be manipulated and reference points within the environment. For instance, the task of straightening a rope on a table can be articulated as pulling one end of the rope to the left side of the center of the table while pulling the other end to the right. Notably, we found that VLMs are highly proficient at generating such spatial relationships when provided with appropriate visual prompts.

Specifically, we first segment all semantic masks from the RGB observation utilizing Segment Anything~\cite{kirillov2023segment}. Next, we perform farthest point sampling (FPS) on these masks to obtain keypoints and reference points, which we then label on the original image. We prompt the VLM to select keypoints and, for each keypoint $k$, to specify a target position $p$ that describes the desired final state of the object set $\mathcal{O}$ upon task completion. The target position $p$ is determined by its spatial offset from a reference point, as provided by the VLM through a set of code assignment statements in the form of $\texttt{p = r + [dx, dx, dz]}$, which denotes that the target position is equal to the reference point $r$'s position added by an offset $\texttt{[dx, dx, dz]}$. We refer to each $(k, p)$ pair as a \textit{target specification}. It is important to note that these target specifications do not have to be strictly satisfied after executing the task, as the vision-language model may occasionally generate infeasible specifications due to its lack of knowledge of object dynamics. Despite this, we found that optimizing for approximate target specifications is generally sufficient to successfully complete the instruction.
For a more detailed prompting process, please refer to the appendix on our project website.




\subsection{Top-K Prompt Library} \label{sec:3.3}

In our experiments, we noticed that providing a few examples of similar tasks significantly improved the performance of the VLM. As a result, we collected a diverse set of examples that cover all the object categories used in our experiments. However, existing VLMs, such as GPT-4V, often have limitations on the number of input images they can task as inputs. Additionally, images consume a substantial number of input tokens, which can be costly and affect the efficiency of the program.


To address this, we developed a prompt retriever that selects optimal examples from the prompt library through score matching. Specifically, we employ CLIP~\cite{radford2021learning} to encode both the input observation $s$ and instruction $\mathcal{L}$, as well as all the examples represented by tuples $(q_i, \text{obs}_i, r_i)$ in the prompt library $\mathcal{P}$, where $q_i$ is the text query, $\text{obs}_i$ is the corresponding observation in the example, and $r_i$ is the response provided by a human expert. We then normalize these latent vectors and obtain similarities by dot product. The matching score $\mathcal{S}_i$ is calculated as a weighted average of the image and text similarities:
\begin{equation}
\mathcal{S}_i = \frac{f_I(s)}{|f_I(s)|} \cdot \frac{f_I(\text{obs}_i)}{|f_I(\text{obs}_i)|} + \lambda \frac{f_T(\mathcal{L})}{|f_T(\mathcal{L})|} \cdot \frac{f_T(q_i)}{|f_T(q_i)|},
\end{equation}
where $f_{I}$ and $f_{T}$ represent the image and text encoders of CLIP, respectively, $\lambda$ is a hyperparameter set to $0.6$. The top $K$ examples are selected and incorporated into our prompt based on the score.



\subsection{Two-Level Closed-Loop Planning} \label{sec:3.4}

In our framework, we propose a two-level closed-loop planning to improve the robustness and effectiveness of the manipulation tasks. 

At the low-level closed loop of model-based planning, once the target specifications are obtained, the next step is to translate them into optimization objectives. We start by projecting the keypoints and their corresponding targets into 3D space. Next, we extract the object points ${o_i}$ from the objects' point clouds and align each keypoint with its nearest object point. The objective is then defined as the sum of the Euclidean distances between keypoints and their corresponding targets:
\begin{equation}
    C(z|\mathcal{L}) = \sum_\text{$i$} \|o_i - p_i\|_2\quad (o_i \in z),
\end{equation}
where $p_i$ represents the 3D target of point $o_i$, and the summation is performed over all target specifications. Using the objective defined in Eqn.~\ref{1} and the dynamics model $f$, we employ the Model Predictive Path Integral (MPPI)~\cite{williams2015model} algorithm, to determine the action to be executed.

However, for specific object categories such as granular pieces, a limited number of keypoints is insufficient to accurately describe the target shape specified by the instruction. Thus, we perform high-level closed-loop re-planning with the VLM, re-prompting it with the current observation and instruction to update target specifications after a series of actions within a loop. This two-level closed-loop planning framework effectively corrects imperfect target specifications and execution errors, etc., ensuring the system's robustness even in the presence of external disturbances.

\section{Experiments}

\begin{figure*}[thbp]
    \centering
    \includegraphics[width=\linewidth]{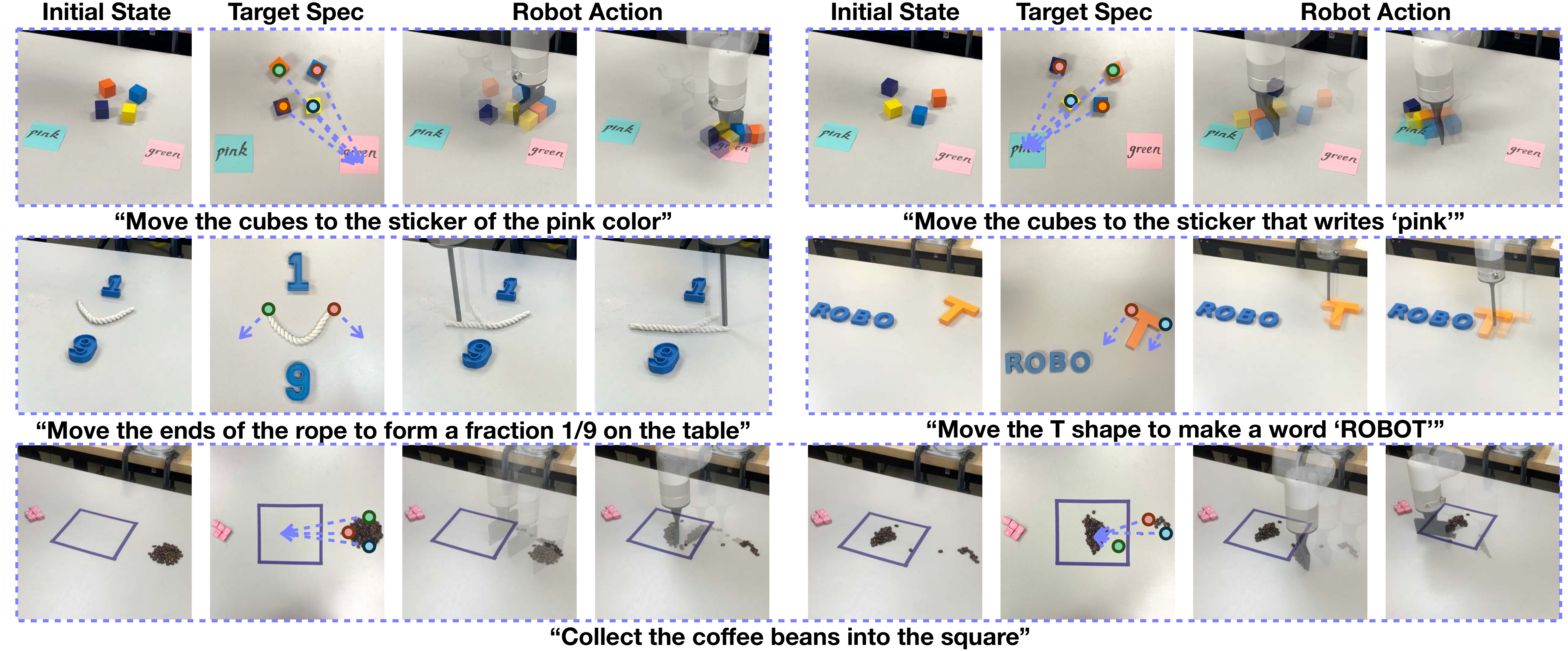}
    \vspace{-15pt}
    \caption{\small \textbf{Qualitative Results of the Rollouts.} We show the target specification and robot executions of various tasks on different objects, highlight the effectiveness of our framework. We show the initial state and the target specification visualization of our system, along with the robot executions, to demonstrate the performance of our framework on various manipulation tasks. Note that we show the granular collection task to exhibit how our VLM-level closed-loop control works in our two VLM-level loops.}
    \vspace{-15pt}

    \label{fig:rollout}
\end{figure*}

The main purpose of our experiments is to verify and analyze the ability of our system to perform a variety of tasks on different objects given various language instructions or experiment configurations. We aim to answer the following research questions:
(1) How well does our system generalize to diverse text instructions and visual scenarios?
(2) How does our framework handle complex manipulation tasks across various object categories?
(3) How does each module contribute to the system's failure cases?

We conduct qualitative evaluations on a diversified set of tasks to demonstrate the effectiveness of our system. To highlight its model-based planning capabilities with neural dynamics models, we compare our framework against two baselines. Additionally, we provide a component-wise experiment error breakdown for a comprehensive analysis of our framework's effectiveness.
Furthermore, as an ablation study, we examine the impact of the hyperparameter $K$ in the top-K prompt library on in-context learning.





To demonstrate the flexibility of our system across various objects, we train the neural dynamics models on 4 different object categories: rope, cubes, granular pieces, and a T-shaped block. The first three categories utilize graph-based neural dynamics models, whereas the T-shaped block employs a state-based neural dynamics model trained using a multilayer perceptron network.


We compare our system with \textbf{MOKA}~\cite{liu2024moka} and \textbf{VoxPoser}~\cite{huang2023voxposer} in a tabletop environment. These two baselines also enable open-vocabulary manipulation and in-context few-shot learning. \textbf{MOKA} builds a framework to prompt VLM to directly generate motion, and \textbf{Voxposer} uses LLM to synthesize 3D voxel map as affordance. To ensure a fair comparison, the prompts and few-shot examples of these 2 systems are adapted to be suitable for our tasks. To avoid the possible overfit in few-shot learning, we ensure that no example in the prompt library is exactly the same as the task. All vision language models and large language models used in our system and two baselines are specified to be GPT-4o.


\subsection{Qualitative Results}

In Fig.~\ref{fig:rollout}, we present 5 tasks featuring different text instructions and visual scenarios, along with their corresponding target specifications generated by our system and the robot’s executions. These examples clearly demonstrate the effectiveness of our framework. In each case, our system generates precise target specifications aligned with the language instructions, and the robot executes the tasks effectively. Notably, two tasks involving cubes start with similar initial configurations but differ in instructions which are easy to confuse. The VLM effectively distinguished semantics and provided precise target specifications, demonstrating its strong visual understanding capabilities.

The coffee bean collection task highlights the benefits of our VLM-level closed-loop planning. Initially, the target specifications were too sparse to fully manipulate the coffee bean pile, leaving a few unspecified beans outside the square after multiple actions. Our system identified these errors and corrected them in the subsequent loop. These examples demonstrate the flexibility of our framework across a wide variety of instructions and environment configurations.

\begin{figure*}[t]
    \centering
    \includegraphics[width=\linewidth]{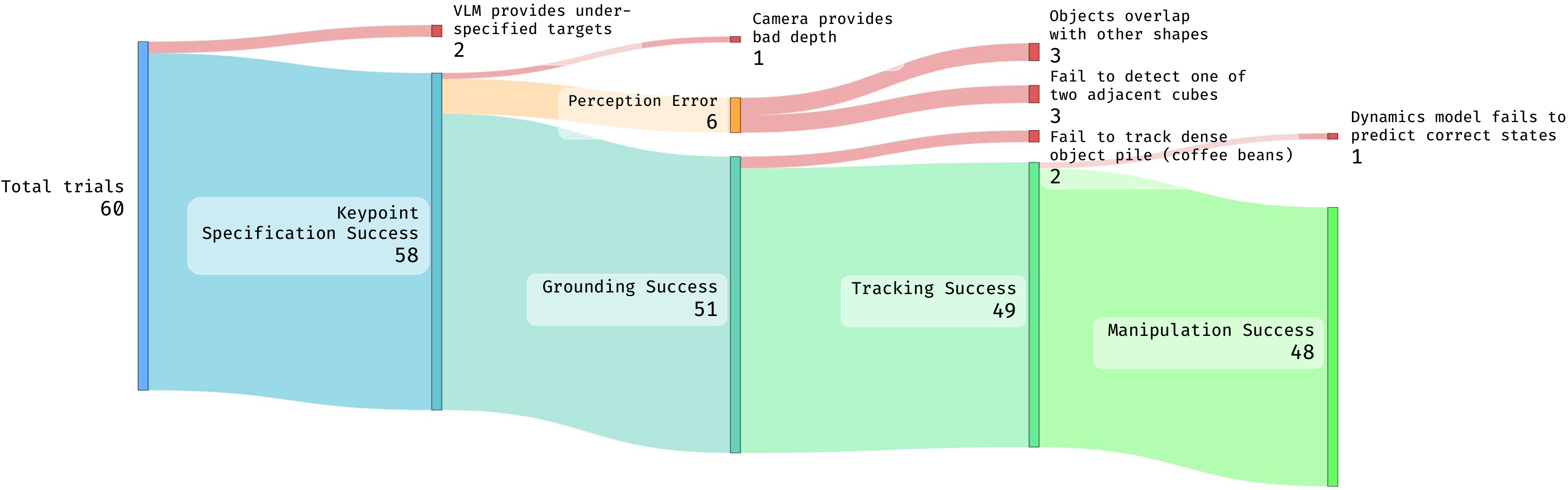}
    \vspace{-15pt}
    \caption{\small \textbf{Visualizations of Error Breakdown.}
    We provide a detailed breakdown of each failure mode, marked in red. While we achieved an $80\%$ success rate across 60 trials for various tasks, the primary cause of failure was perception errors, accounting for $10\%$ of all trials and $50\%$ of the failure cases.}
    \vspace{-15pt}
    \label{fig:error}
\end{figure*}

\subsection{Quantitative Results}


We compare our system with two baselines on a total of six tasks across the four object categories. The results are shown in Tab.~\ref{table:quanti}. For each task, we evaluate the success rate by measuring the Chamfer distance between the point clouds of the objects after a specified number of robot actions are applied and the corresponding target point clouds. The quantitative results are shown as the total number of successful trials out of a total of 10. The text instructions for all evaluation tasks are listed in Tab.~\ref{tab:instruction}.

\begin{table}[htbp]
\centering
\begin{tabular}{lccc}
    \toprule
    Methods & MOKA~\cite{liu2024moka} & VoxPoser~\cite{huang2023voxposer} & \textbf{Ours}\\
    \midrule
    Rope Straightening & 2/10 & 0/10 & \textbf{8/10}\\
    \midrule
    Cube Collection & 0/10 & 3/10 & \textbf{6/10}\\
    Cube Movement & 6/10 & 3/10 & \textbf{10/10}\\
    \midrule
    Granular Collection & 0/10 & 1/10 & \textbf{10/10}\\
    Granular Movement & 0/10 & 1/10 & \textbf{6/10}\\
    \midrule
    T Movement & 0/10 & 0/10 & \textbf{8/10}\\
    \midrule
    \textbf{Total} & 13.3\% & 13.3\% & \textbf{80.0}\%\\
    \bottomrule
\end{tabular}
\vspace{-5pt}
\caption{\small \textbf{Quantitative results of our evaluation.} Our method achieved relatively high performance across all evaluation tasks compared to the two baselines, while the failures in Cube Collection and Granular Movement were primarily caused by perception.}
\vspace{-10pt}
\label{table:quanti}
\end{table}

As demonstrated in the quantitative evaluation results, our system is superior to these existing methods by a large margin on the evaluation tasks. The main reason for such a performance gap is that the existing methods ignore the fine-grained representation of objects or actions and lack the knowledge of object dynamics, which leads to their limited capability to manipulate deformable objects or rigid objects of more complex shapes. In contrast, our system utilizes the keypoint-based representation for visual prompting and dynamics learning. This enables our system to predict the future state of those complex objects and perform model-based planning for manipulation.

\begin{table}[htbp]
    \centering
    \begin{tabular}{c|l}
        \hline
        Rope Straightening & ``Straighten the rope.''\\
        \hline
        Cube Collection & ``Move all the cubes to the pink cross.''\\
        \hline
        Cube Movement & ``Move the yellow cube to the red cross.''\\
        \hline
        Granular Collection & ``Collect all the coffee beans together.''\\
        \hline
        Granular Movement & ``Move all the coffee beans to the red cross.''\\
        \hline
        T Movement & ``Move the orange T into the pink square.''\\
        \hline
    \end{tabular}
    \caption{\small The input instructions of each evaluation task.}
    \vspace{-10pt}
    \label{tab:instruction}
\end{table}

\subsection{Error Breakdown}


We conducted a manual analysis of the failure cases encountered during our experiments, as shown in Fig.~\ref{fig:error}. As illustrated, 1) the perception module accounted for the majority of errors. This includes instances where the module failed to detect objects, particularly when they overlapped with other shapes, such as a cross sign on the table, or failed to distinguish between two adjacent cubes. 2) The second most significant source of error stemmed from the target specification and the tracking module. Errors in target specification typically arose from the VLM providing under-specified targets, i.e., an insufficient number of target specifications to complete the task. The tracking module commonly failed when tracking keypoints in dense object piles, such as coffee beans. 3) Additionally, a smaller proportion of failures were attributed to the dynamics model and hardware, where the dynamics model occasionally produced inaccurate predictions, and the RGBD camera sometimes provided inaccurate depth values.

\subsection{Ablation of Top-K Prompt Library}


\begin{table}[htbp]
\vspace{-10pt}
    \centering
    \begin{tabular}{c|c|c|c|c|c}
    \toprule
 & Category & Top-0 & Top-1 & Top-3 & Top-5 \\
    \midrule
    Success Rate & 10/10 & 2/10 & 3/10 & 10/10 & 7/10 \\
    \bottomrule
    \end{tabular}
    \caption{\small The quantitative results of different $K$ values.}\vspace{-10pt}
    \label{tab:ablation}
\end{table}


In this ablation, we evaluate the success rate on the Rope Straightening task for different $K$ values, as well as a special prompting method labeled ``category'' where examples are manually selected by human expert. The results are presented in Tab.~\ref{tab:ablation}. Notably, when $K = 3$, the prompt retriever achieves performance on par with that of a human expert. However, increasing $K$ introduces less relevant examples, which lowers the overall prompt quality and results in a reduced success rate for prompt retriever when $K=5$.


\section{Conclusion \& Limitations}


In this work, we propose a novel open-vocabulary robotic manipulation system KUDA, which unifies the visual prompting of the vision language models and dynamics learning through keypoint-based representation. Utilizing this flexible representation, KUDA leverages the vision language model to deal with various high-level human language instructions, and in the meantime utilizes model-based planning with dynamics models to generate robot actions, to perform complex manipulation tasks on various object categories, with different language instructions. Our experiments demonstrate the effectiveness and versatility of our framework.

However, KUDA also has several limitations: First, it uses a top camera to capture visual observations for the vision language model, limiting its ability to perform tasks with more complex 3D spatial relationships. Second, the dynamics models in our work are trained in simulations, leading to an inevitable sim-to-real gap and limited generalization to different object categories. We believe that with the development of related fields in the future, the problems above will be finally eliminated. We hope that KUDA will inspire more future work on incorporating knowledge of dynamics into more versatile robotic systems.

\section{Acknowledgement}

This work is partially supported by the Toyota Research Institute (TRI), the Sony Group Corporation, and Google. This article solely reflects the opinions and conclusions of its authors and should not be interpreted as necessarily representing the official policies, either expressed or implied, of the sponsors.

\addtolength{\textheight}{-0cm}   





\bibliographystyle{IEEEtran}
\bibliography{IEEEabrv,reference}

\newpage

\end{document}